\documentclass[conference,a4paper]{IEEEtran}
\IEEEoverridecommandlockouts
\usepackage[hidelinks]{hyperref}
\usepackage{xurl}  % allows \url to break at any character
\usepackage[cmex10]{amsmath}%American Math Society(AMS) math formatting
\usepackage{amssymb,amsfonts}%AMS extra symbols and fonts
\usepackage[most]{tcolorbox}
\usepackage{amsmath,amssymb,amsfonts}
\usepackage{algorithmic}
\usepackage{graphicx}
\usepackage{textcomp}
\usepackage{xcolor}
\usepackage{mdframed}
\usepackage{tcolorbox}
\usepackage{enumitem}
\usepackage{array}
\usepackage{colortbl}
\usepackage{dblfloatfix}%fix double column figure ordering and placement
\usepackage[ruled,vlined]{algorithm2e}
\usepackage{booktabs}
\usepackage{siunitx}
\usepackage[numbers,compress]{natbib}
\usepackage{texnames}
\usepackage{bm,bbm}
\usepackage{orcidlink}
\usepackage{float}
\begin{document}

\title{WildfireVLM: AI-powered Analysis for Early Wildfire Detection and Risk Assessment Using Satellite Imagery }

\author{
    \IEEEauthorblockN{Aydin Ayanzadeh\orcidlink{0000-0002-8816-3204}, 
                      Prakhar Dixit\orcidlink{0000-0003-0450-5420}, 
                      Sadia Kamal\orcidlink{0009-0004-7679-465X}, 
                      and  Milton Halem\orcidlink{0000-0002-5614-3612}}
    \IEEEauthorblockA{\textit{Department of Computer Science and Electrical Engineering}\\
        \textit{University of Maryland, Baltimore County}\\
        Baltimore, MD 21250, USA\\
        \{aydina1, pdixit1, sadia1402, halem\}@umbc.edu}
}
\maketitle

\begin{abstract}

Wildfires are a growing threat to ecosystems, human lives, and infrastructure, with their frequency and intensity rising due to climate change and human activities. Early detection is critical, yet satellite-based monitoring remains challenging due to faint smoke signals, dynamic weather conditions, and the need for real-time analysis over large areas. We introduce WildfireVLM,  an AI framework that combines satellite imagery wildfire detection with language-driven risk assessment. We construct a labeled wildfire and smoke dataset using imagery from Landsat-8/9, GOES-16, and other publicly available Earth observation sources, including harmonized products with aligned spectral bands. WildfireVLM employs YOLOv12 to detect fire zones and smoke plumes, leveraging its ability to detect small, complex patterns in satellite imagery. We integrate Multimodal Large Language Models (MLLMs) that convert detection outputs into contextualized risk assessments and prioritized response recommendations for disaster management. We validate the quality of risk reasoning using an LLM-as-judge evaluation with a shared rubric. The system is deployed using a service-oriented architecture that supports real-time processing, visual risk dashboards, and long-term wildfire tracking, demonstrating the value of combining computer vision with language-based reasoning for scalable wildfire monitoring. The code and dataset are publicly available on GitHub\footnote{\url{https://github.com/Ayanzadeh93/_WildfireVLM_}}.
\end{abstract}


\section{Introduction}
\begin{figure}[!t]
    \centering
    \includegraphics[width=\columnwidth]{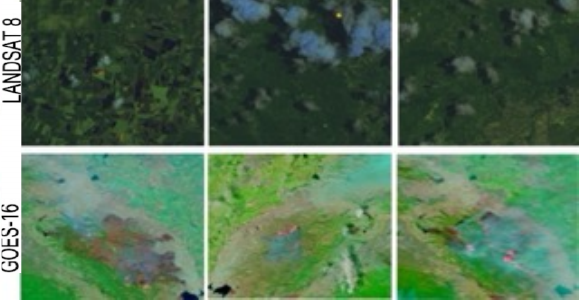}
    \caption{Demonstration of GOES-16 and Landsat 8 dataset.}
    \label{fig:dataset}
\end{figure}

\begin{figure*}[!t]
    \centering
    \includegraphics[width=0.88\textwidth, height=6.2cm]{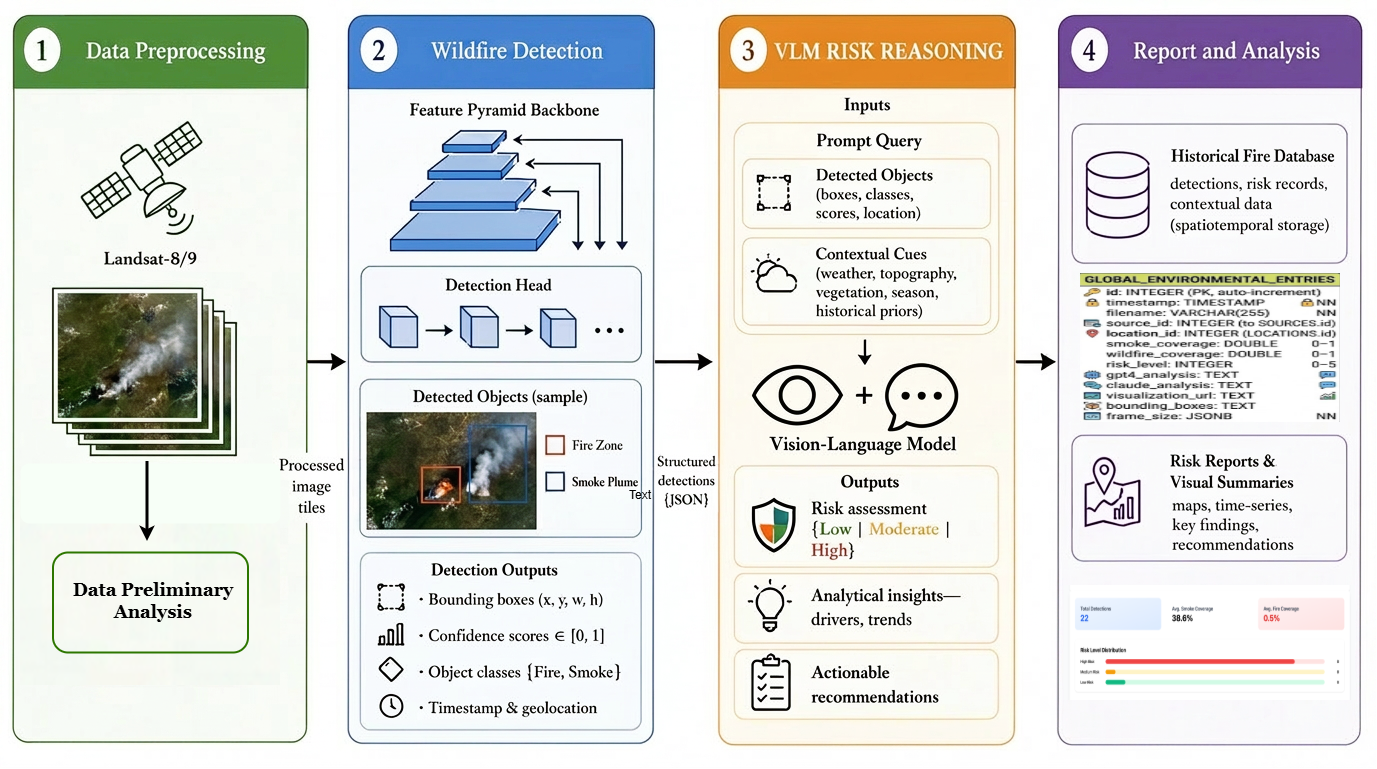}
    
    \caption{System architecture of WildfireVLM, consisting of four integrated modules: (1) Input module processes Landsat-8/9 satellite imagery via preprocessing; (2) YOLOv12 Detection Core performs object detection to identify fire zones and smoke plumes, outputting bounding boxes, confidence scores, and classes; (3) Vision-Language Risk Reasoning module uses a language model for contextual understanding, generating risk assessments, insights, and recommendations; and (4) Risk and Output module stores results in a historical fire database, supports temporal analysis, and generates risk reports with visual summaries for decision support.}
    \label{fig:yolov12_arch}
\end{figure*}

Wildfires are among the most destructive natural disasters, with their frequency and severity increasing due to climate change and human activities \cite{xofis2020fire}. Each year, wildfires devastate millions of acres of forest, threaten ecosystems and human lives, degrade air quality, and impose economic losses amounting to tens of billions of USD \cite{wang2021economic,xu2021forest}. Regions such as California, Australia, the Amazon rainforest, and the Mediterranean basin have experienced increasingly large-scale wildfire events driven by prolonged droughts, rising temperatures, and land-use changes \cite{halofsky2020changing,hessburg2022climate,westerling2008climate,goss2020climate,blanchi2010meteorological}. These trends underscore the urgent need for early wildfire detection and effective risk assessment. Traditional wildfire monitoring approaches often rely on manual observation, sparse sensor networks, or static risk models, which can delay detection and limit situational awareness. In contrast, the growing availability of satellite imagery, combined with advances in artificial intelligence (AI), enables large-scale and continuous wildfire monitoring \cite{liu2025advancements}. Smoke detection is particularly important for early warning, as smoke frequently appears before visible flames. Deep learning–based object detection models, such as YOLO (You Only Look Once), have demonstrated strong performance in detecting wildfire features, including smoke plumes and fire fronts \cite{redmon2016you}.

Despite these advances, accurate wildfire detection in satellite imagery remains challenging due to lower spatial resolution, atmospheric interference, cloud cover, and visually ambiguous backgrounds.\cite{negash2025review} Moreover, most existing methods focus primarily on visual detection or classification and provide limited higher-level interpretation or risk assessment. To address these limitations, we introduce \textit{WildfireVLM}, an AI-driven framework for satellite-based wildfire and smoke detection with integrated risk reasoning.\cite{jin2025smoke,fernandes2022automatic} WildfireVLM combines YOLO-based object detection with vision-language models (VLMs) to generate contextual risk assessments, heatmaps, and actionable recommendations, moving beyond detection toward comprehensive decision support for disaster management. Unlike WildfireGPT~\cite{xie2024wildfiregpt}, which employs a text-only LLM agent with retrieval augmentation, WildfireVLM integrates visual object detection outputs with multimodal language models for end-to-end satellite-based risk assessment.

\begin{figure*}[!t]
    \centering
    \scalebox{1}[0.9]{\includegraphics[width=0.88\textwidth]{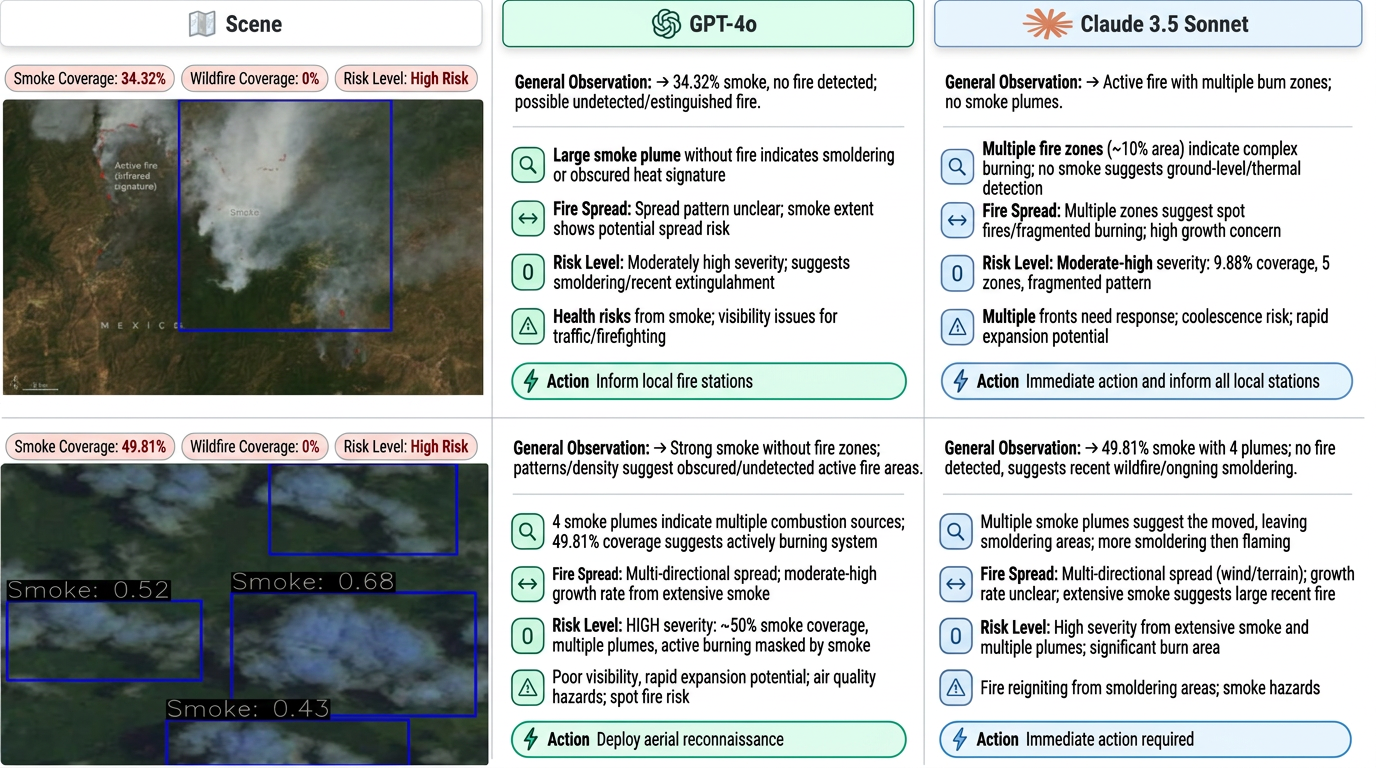}}
    \caption{Wildfire risk assessment comparison between GPT-4o and Claude 3.5 Sonnet on the proposed dataset. Left: Demonstration of wildfire detection. Right: LLM-generated structured risk assessments including general observations, fire behavior analysis, spread potential, severity classification, and actionable recommendations/ Insight Action.}
    \label{fig:output}
\end{figure*}

\section{Related Work}

Recent advances in computer vision have significantly improved wildfire and smoke detection, with YOLO-based object detection models emerging as a dominant approach due to their real-time performance and strong localization accuracy. Gonçalves \cite{goncalves2024yolo} demonstrated the effectiveness of YOLO architectures for wildfire and smoke detection across ground-based and aerial imagery, while Kim and Muminov \cite{kim2023smoke} applied YOLOv7 to UAV imagery for early-stage fire and smoke monitoring over large areas. Complementary to detection-focused pipelines, Navardi et al. \cite{navardi2022toward} demonstrated that object-detector-extracted features can serve as robust inputs to reinforcement learning policies for autonomous drone avigation, offering a sim-to-real transfer strategy applicable to aerial monitoring platforms. Maillard \cite{maillard2024yolo} introduced YOLO-NAS, a neural architecture search approach that balances detection accuracy and efficiency for large-scale monitoring, while Yang \cite{yang2023improved} improved YOLOv5 for aerial smoke detection via transfer learning to address scale variation, viewpoint changes, and environmental noise. More recent lightweight variants target improved sensitivity to fire structures; Tao et al. \cite{tao2025improved} proposed YOLO11-RLN with attention-based fire-line modeling, and Alkhammash \cite{alkhammash2025multi} explored YOLOv11-based multi-class fire classification for close-range fire-type recognition. 

Despite these advances, most existing methods rely on ground-level or UAV imagery and primarily focus on visual detection or classification outputs. Detecting subtle fire and smoke patterns in satellite imagery remains challenging due to lower spatial resolution and atmospheric interference, and prior work largely lacks higher-level contextual interpretation or risk assessment.\cite{elhanashi2025early,yin2026bgc} In contrast, this work targets satellite-based wildfire detection and integrates vision–language models to enable contextual risk reasoning, moving beyond object detection toward comprehensive decision support. Vision-language models have been adapted for remote sensing through grounded spatial reasoning~\cite{kuckreja2024geochat}, captioning benchmarks~\cite{hu2023rsgpt}, and multi-sensor comprehension~\cite{zhang2024earthgpt}. Beyond remote sensing, the application of LLMs to structured spatial tasks, such as indoor navigation, has demonstrated robustness~\cite{ayanzadeh2024floorplan2guide}. Similarly, we leverage language-based reasoning to interpret satellite-based wildfire detection for risk assessment. While LLM-as-a-judge evaluation~\cite{zheng2023judging,li2024llm_judges_survey} has emerged as a scalable alternative to human assessment, we specifically target contextual risk reasoning for disaster management.

\section{Data Collection}

% \subsection{Data Sources}
The dataset used in this study was compiled from three primary sources: \textbf{Landsat 8}, \textbf{GOES-16}, and other publicly available \textbf{NASA Earth observation imagery}. These sources were selected because they provide complementary spatial, temporal, and spectral information for wildfire monitoring. Landsat 8 offers multispectral and thermal observations at moderate spatial resolution, while GOES-16 provides high-temporal-frequency imagery for monitoring wildfire progression and smoke dispersion over the Americas. In addition, other NASA Earth observation imagery was incorporated to increase variability in scene content and improve coverage of wildfire and smoke events across different conditions.

\subsection{Data Preprocessing}
The collected imagery covers wildfire-prone geographic regions and includes samples from multiple time periods to capture variation in fire intensity, smoke spread, and background land cover. During preprocessing, corrupted samples, duplicate images, low-quality scenes, and irrelevant images without visible wildfire or smoke content were removed to improve dataset quality. Each image was then cropped into \(416 \times 416\) pixel patches to create a uniform input format for model training and evaluation.
%Annotation was performed manually using the \textbf{LabelImg} image annotation tool. For each image, the region of interest was marked and assigned to one of two classes: \textbf{Wildfire} and \textbf{Smoke}.
\subsection{Data Annotation}
Data annotation process enabled supervised training of the object detection models by explicitly labeling wildfire and smoke instances in satellite imagery. The developed dataset contains 3,771 images, which were divided into training (70\%), validation (15\%), and test (15\%) subsets.

\textbf{Landsat 8:} Landsat 8, as depicted in Figure~\ref{fig:dataset}, provides multispectral imagery suitable for environmental monitoring. Equipped with the Operational Land Imager (OLI) and Thermal Infrared Sensor (TIRS), it offers 15--30~m spatial resolution products for analyzing wildfire-affected regions. Landsat 8 follows a 16-day revisit cycle and provides multispectral imagery that supports both visual and thermal characterization of active fires and smoke-related changes.

\textbf{GOES-16:} GOES-16 is a geostationary environmental satellite that provides high-frequency imagery (e.g., every 5 minutes) across 16 spectral bands at approximately 2~km spatial resolution. Its rapid refresh rate enables real-time monitoring of wildfire evolution and smoke transport. GOES-16 is positioned in geostationary orbit and provides continuous coverage over the Americas.

The data preprocessing stage involved removing corrupted or irrelevant samples, resizing images to a uniform resolution of $416\times416$ pixels, and annotating imagery into two classes: wildfire and smoke. The dataset comprises 3{,}771 images and is split into training (70\%), validation (15\%), and test (15\%) sets. We adopted a service-oriented architecture to support scalable wildfire detection and risk assessment.
\section{System Architecture}

The system integrates object detection models and multimodal large language models (MLLMs) through an API-driven design to ensure modularity, reliability, and extensibility. The workflow begins at the client layer, which provides a web-based interface for data submission, visualization, and result inspection. The application layer consists of two primary components: a detection core and a database manager, which is depicted in Figure~\ref{fig:yolov12_arch}. 
%\subsection{Object Detection} 
%\subsection{Language Model} 

The Detection Core interfaces with multiple object detection models, including  YOLOv12~\cite{tian2025yolov12}, YOLOv11~\cite{yolov11}, YOLOv8~\cite{yolov8}, and YOLO-NAS~\cite{yolonas}. The YOLOv12 object detection architecture identifies wildfire-related features, such as smoke plumes and active fire zones, in satellite imagery. The Database Manager interacts with a PostgreSQL backend that stores satellite images, detection metadata, temporal information, and derived numerical measurements, such as wildfire and smoke coverage, and generates responses using employed language models.

Risk assessment is performed by external multimodal language models, including GPT-4o\cite{gpt4o} and Claude 3.5 Sonnet, which receive object detection outputs as structured auxiliary inputs. Incorporating detection results into the language models improves robustness and interpretability in wildfire risk analysis. Asynchronous processing is employed throughout the pipeline to handle concurrent requests while maintaining responsive user interaction, perform higher-level risk assessment, and generate structured analytical reports. By combining visual detection results with language-based reasoning, WildfireVLM enables an end-to-end pipeline for real-time wildfire detection and risk interpretation.

\begin{table}[!t]
\renewcommand{\arraystretch}{1.3}
\caption{Comparative Analysis of YOLO Variants on Satellite Imagery Dataset}
\centering
\begin{tabular}{|c|c|c|c|c|}
\hline
\textbf{Model} & \multicolumn{4}{|c|}{\textbf{Performance Metrics}} \\
\cline{2-5}
 & \textbf{mAP (\%)} & \textbf{Prec. (\%)} & \textbf{Rec. (\%)} & \textbf{F1-score (\%)} \\
\hline
YOLOv8  & 72.1 & 60.7 & 67.6 & 64.0 \\
YOLOv11 & \textbf{84.1} & 51.7 & \textbf{89.8} & 65.6 \\
YOLO-NAS & 54.1 & 56.0 & 57.1 & 56.6 \\
YOLOv12 & 74.7 & \textbf{81.1} & 74.8 & \textbf{77.8} \\
\hline
\multicolumn{5}{l}{$^{\mathrm{a}}$Results based on evaluation of the satellite imagery dataset.}
\end{tabular}
\label{tab:model_comparison}
\end{table}

\section{Experimental Results}

We employed GPT-4o and Claude 3.5 Sonnet for the risk reasoning task. Both models were used with a temperature of 0.2, a maximum generation length of 1024 tokens, and top-$p$ sampling with $ p=0.95$. The system uses a FastAPI backend with a PostgreSQL database for metadata and history management. For object detection training, standard data augmentation techniques (random scaling, rotation, and brightness adjustment) were applied to improve generalization. Detection performance is summarized using precision, recall, F1-score, and mean Average Precision (mAP). We report mAP@50 computed at an Intersection-over-Union (IoU) threshold of 0.50. For deployment in our service-oriented system, inference post-processing uses a confidence threshold of 40 and an overlap/NMS setting of 30, enabling consistent operational behavior across model variants.

\subsection{Wildfire Detection}

Quantitative evaluation results are summarized in Table~\ref{tab:model_comparison}, which compares the performance of YOLOv8, YOLOv11, YOLO-NAS, and YOLOv12 on the satellite imagery dataset. YOLOv12 achieves the highest precision (81.1\%) and the best F1-score (77.8\%), making it the superior choice for minimizing false positives and providing reliable risk reasoning. While YOLOv11 achieves higher raw mAP (84.1\%) and recall (89.8\%), its lower precision (51.7\%) can lead to over-reporting. Consequently, YOLOv12 is selected as the primary detection core for WildfireVLM due to its robust performance and balanced detection profile, which is critical for operational disaster management. Figure~\ref{fig:output} demonstrates the system's output, showing YOLOv12 detection results alongside LLM-generated risk assessments.

\begin{figure}[!ht]
\begin{tcolorbox}[
    enhanced,
    title=Wildfire Analysis Prompt,
    colback=blue!3!white,
    colframe=blue!60!black,
    fonttitle=\bfseries\small,
    coltitle=white,
    boxrule=0.8pt,
    arc=3pt,
    left=5pt,right=5pt,top=5pt,bottom=5pt]
\small
\textbf{Role:} You are a senior wildfire analyst specializing in satellite imagery analysis. Analyze the following detection data from
satellite imagery and provide a comprehensive risk assessment.
In the first step, you need to analyze the image and provide a detailed analysis addressing the following points,
There might be some bounding boxes of ROI in the image, but it is not guaranteed that all the bounding boxes are
detected or localized correctly.
** If you find a wildfire that the model did not detect, you can assume that the wildfire is detected and provide the
analysis based on your point. 

\textbf{Input Parameters:}
Image size, smoke coverage (\%), wildfire coverage (\%)

\textbf{Analysis Requirements:}
\begin{enumerate}[leftmargin=*,itemsep=0pt,parsep=0pt]
\item Visual assessment independent of bounding boxes
\item Fire behavior from smoke patterns and burn distribution
\item Spread potential evaluation (pattern, growth rate)
\item Severity classification from coverage metrics
\item Critical risk identification
\item Actionable recommendations / Insight 
\end{enumerate}

\textbf{Key Considerations:} Plume density, fire zone clustering, spatial relationships, and infrastructure impact.
\end{tcolorbox}
\label{box:prompt}
\end{figure}

\subsection{Language Model Evaluation}
To evaluate language model quality in wildfire risk analysis, we provide both models with structured prompts that specify detection parameters and analysis requirements. The complete prompt structure is provided in the Wildfire Analysis Prompt box. We generate risk assessment responses for the test set using GPT-4o and Claude Sonnet 3.5. Following the LLM-as-a-judge paradigm~\cite{zheng2023judging}, we then employ Claude Sonnet 4.5 as an external LLM-based judge to assess the outputs of both models under a shared, rubric-based evaluation framework. The external LLM evaluates semantic correctness, risk reasoning, and actionable clarity, assigning a score of 1-10 to each response. The LLM-as-judge evaluation yields mean scores of \textbf{7.03 }for GPT-4o and \textbf{6.16} for Claude 3.5 Sonnet, indicating that GPT-4o performs better at generating comprehensive risk assessments with higher semantic accuracy and greater actionable clarity. Both models provide reliable risk analysis capabilities, though GPT-4o aligns more consistently with expert-level wildfire assessment criteria.

\section{Conclusion}

In this study, we introduced WildfireVLM, an AI-driven framework for early wildfire detection and risk assessment using satellite imagery. WildfireVLM achieves 81.1\% precision and a 77.8\% F1-score on the Landsat–GOES wildfire dataset. Our results demonstrate the potential of AI and computer vision to enhance wildfire monitoring and risk mitigation. Future work will focus on improving model performance and expanding the dataset to cover additional regions and environmental conditions. In addition, we plan to fine-tune domain-specific LLMs on expert wildfire assessments annotated by human experts to improve contextual reasoning and to explore the capability of language models to assess risk levels and detect wildfires at earlier stages, thereby enhancing disaster management across diverse wildfire scenarios.

\bibliographystyle{IEEEtranN}
\bibliography{references}

@article{alkhammash2025multi,
  title={Multi-classification using YOLOv11 and hybrid YOLO11n-MobileNet models: A fire classes case study},
  author={Alkhammash, Eman H},
  journal={Fire},
  volume={8},
  number={1},
  pages={17},
  year={2025},
  publisher={MDPI}
}

@article{tao2025improved,
  title={Improved Lightweight YOLOv11 Algorithm for Real-Time Forest Fire Detection},
  author={Tao, Ye and Li, Bangyu and Li, Peiru and Qian, Jin and Qi, Liang},
  journal={Electronics},
  volume={14},
  number={8},
  pages={1508},
  year={2025},
  publisher={MDPI}
}

@article{xofis2020fire,
  author  = {P. Xofis and G. Tsiourlis and P. Konstantinidis},
  title   = {A fire danger index for the early detection of areas vulnerable to wildfires in the eastern {Mediterranean} region},
  journal = {Euro-Mediterranean Journal for Environmental Integration},
  year    = {2020},
  volume  = {5},
  number  = {2},
  pages   = {32}
}

@article{wang2021economic,
  author  = {D. Wang and D. Guan and S. Zhu and M. M. Kinnon and G. Geng and Q. Zhang and H. Zheng and T. Lei and S. Shao and P. Gong and others},
  title   = {Economic footprint of {California} wildfires in 2018},
  journal = {Nature Sustainability},
  year    = {2021},
  volume  = {4},
  number  = {3},
  pages   = {252--260}
}

@article{xu2021forest,
  author  = {R. Xu and H. Lin and K. Lu and L. Cao and Y. Liu},
  title   = {A forest fire detection system based on ensemble learning},
  journal = {Forests},
  year    = {2021},
  volume  = {12},
  number  = {2},
  pages   = {217}
}

@article{halofsky2020changing,
  author  = {J. E. Halofsky and D. L. Peterson and B. J. Harvey},
  title   = {Changing wildfire, changing forests: the effects of climate change on fire regimes and vegetation in the {Pacific Northwest, USA}},
  journal = {Fire Ecology},
  year    = {2020},
  volume  = {16},
  number  = {1},
  pages   = {1--26}
}

@article{hessburg2022climate,
  author  = {P. F. Hessburg and S. Charnley and A. N. Gray and T. A. Spies and D. W. Peterson and R. L. Flitcroft and K. L. Wendel and J. E. Halofsky and E. M. White and J. Marshall},
  title   = {Climate and wildfire adaptation of inland northwest {US} forests},
  journal = {Frontiers in Ecology and the Environment},
  year    = {2022},
  volume  = {20},
  number  = {1},
  pages   = {40--48}
}

@article{westerling2008climate,
  author  = {A. L. Westerling and B. P. Bryant},
  title   = {Climate change and wildfire in {California}},
  journal = {Climatic Change},
  year    = {2008},
  volume  = {87},
  pages   = {231--249}
}

@article{tian2025yolov12,
  title={Yolov12: Attention-centric real-time object detectors},
  author={Tian, Yunjie and Ye, Qixiang and Doermann, David},
  journal={arXiv preprint arXiv:2502.12524},
  year={2025}
}

@article{elhanashi2025early,
  title={Early fire and smoke detection using deep learning: A comprehensive review of models, datasets, and challenges},
  author={Elhanashi, Abdussalam and Essahraui, Siham and Dini, Pierpaolo and Saponara, Sergio},
  journal={Applied Sciences},
  volume={15},
  number={18},
  pages={10255},
  year={2025},
  publisher={MDPI}
}

@article{yin2026bgc,
  title={BGC-LiteNet: BeiDou grid code embedded lightweight neural architecture for real-time UAV fire detection and localization},
  author={Yin, Haiwen and Yu, Yong and Hong, Andong and Hu, Mengyang and Wang, Shunhang and Zhang, Zhifan},
  journal={Scientific Reports},
  year={2026},
  publisher={Nature Publishing Group UK London}
}

@article{jin2025smoke,
  title={From smoke to fire: A forest fire early warning and risk assessment model fusing multimodal data},
  author={Jin, Peixian and Cheng, Pengle and Liu, Xiaodong and Huang, Ying},
  journal={Engineering Applications of Artificial Intelligence},
  volume={152},
  pages={110848},
  year={2025},
  publisher={Elsevier}
}

@article{fernandes2022automatic,
  title={Automatic early detection of wildfire smoke with visible light cameras using deep learning and visual explanation},
  author={Fernandes, Armando M and Utkin, Andrei B and Chaves, Paulo},
  journal={IEEE Access},
  volume={10},
  pages={12814--12828},
  year={2022},
  publisher={IEEE}
}

@inproceedings{negash2025review,
  title={Review of wildfire detection, fighting, and technologies: Future prospects and insights},
  author={Negash, Natnael M and Sun, Liang and Fan, Chao and Shi, Di and Wang, Fengyu},
  booktitle={AIAA AVIATION FORUM AND ASCEND 2025},
  pages={3469},
  year={2025}
}

@article{liu2025advancements,
  title={Advancements in artificial intelligence applications for forest fire prediction},
  author={Liu, Hui and Shu, Lifu and Liu, Xiaodong and Cheng, Pengle and Wang, Mingyu and Huang, Ying},
  journal={Forests},
  volume={16},
  number={4},
  pages={704},
  year={2025},
  publisher={MDPI}
}

@article{navardi2022toward,
  title={Toward real-world implementation of deep reinforcement learning for vision-based autonomous drone navigation with mission},
  author={Navardi, Mozhgan and Dixit, Prakhar and Manjunath, Tejaswini and Waytowich, Nicholas R and Mohsenin, Tinoosh and Oates, Tim},
  journal={arXiv preprint arXiv:2208.06456},
  year={2022}
}

@article{goss2020climate,
  author  = {M. Goss and D. L. Swain and J. T. Abatzoglou and A. Sarhadi and C. A. Kolden and A. P. Williams and N. S. Diffenbaugh},
  title   = {Climate change is increasing the likelihood of extreme autumn wildfire conditions across {California}},
  journal = {Environmental Research Letters},
  year    = {2020},
  volume  = {15},
  number  = {9},
  pages   = {094016}
}

@article{blanchi2010meteorological,
  author  = {R. Blanchi and C. Lucas and J. Leonard and K. Finkele},
  title   = {Meteorological conditions and wildfire-related houseloss in {Australia}},
  journal = {International Journal of Wildland Fire},
  year    = {2010},
  volume  = {19},
  number  = {7},
  pages   = {914--926}
}

@inproceedings{redmon2016you,
  author    = {J. Redmon and S. Divvala and R. Girshick and A. Farhadi},
  title     = {You only look once: Unified, real-time object detection},
  booktitle = {Proceedings of the IEEE Conference on Computer Vision and Pattern Recognition},
  year      = {2016},
  pages     = {779--788}
}

@article{goncalves2024yolo,
  author  = {L. A. O. Gon{\c{c}}alves and R. Ghali and M. A. Akhloufi},
  title   = {{YOLO}-based models for smoke and wildfire detection in ground and aerial images},
  journal = {Fire},
  year    = {2024},
  volume  = {7},
  number  = {4},
  pages   = {140},
  url     = {https://www.mdpi.com/2571-6255/7/4/140}
}

@article{kim2023smoke,
  author  = {B. Kim and N. Muminov},
  title   = {Smoke detection in {UAV} images using {YOLOv7}},
  journal = {Sensors},
  year    = {2023},
  volume  = {23},
  number  = {15},
  pages   = {6701}
}

@inproceedings{maillard2024yolo,
  author    = {A. Maillard and others},
  title     = {Wildfire and smoke detection using {YOLO-NAS}},
  booktitle = {IEEE Conference Proceedings},
  year      = {2024},
  url       = {https://ieeexplore.ieee.org/document/10585773}
}

@article{yang2023improved,
  author  = {Yang and others},
  title   = {Improved {YOLOv5} for aerial smoke detection},
  journal = {Fire Technology},
  year    = {2023},
  volume  = {59},
  pages   = {1--20}
}

@misc{yolov11,
  author       = {G. Jocher and others},
  title        = {{YOLOv11}: State-of-the-art object detection},
  year         = {2024},
  howpublished = {Ultralytics},
  url          = {https://github.com/ultralytics/ultralytics}
}

@misc{yolov8,
  author       = {G. Jocher and A. Chaurasia and J. Qiu},
  title        = {{YOLOv8}: A real-time object detection system},
  year         = {2023},
  howpublished = {arXiv preprint arXiv:2305.09972}
}

@misc{yolonas,
  author       = {Deci AI},
  title        = {{YOLO-NAS}: Neural architecture search for object detection},
  year         = {2023},
  howpublished = {Technical Report},
  url          = {https://deci.ai/blog/yolo-nas-object-detection-foundation-model/}
}

@misc{gpt4o,
  title        = {GPT-4 Technical Report},
  author       = {{OpenAI}},
  year         = {2023},
  month        = mar,
  howpublished = {OpenAI Research},
  url          = {https://openai.com/index/gpt-4-research/},
  note         = {Published 2023.}
}

@inproceedings{kuckreja2024geochat,
  title={GeoChat: Grounded Large Vision-Language Model for Remote Sensing},
  author={Kuckreja, Kartik and Danish, Muhammad Saad and Naseer, Muzammal and Das, Abhijit and Khan, Salman and Khan, Fahad Shahbaz},
  booktitle={Proc. IEEE/CVF Conference on Computer Vision and Pattern Recognition (CVPR)},
  pages={27831--27840},
  year={2024}
}

@article{hu2023rsgpt,
  title={{RSGPT}: A Remote Sensing Vision Language Model and Benchmark},
  author={Hu, Yuan and Yuan, Jianlong and Wen, Congcong and Lu, Xiaonan and Xian, Yutong},
  journal={arXiv preprint arXiv:2307.15266},
  year={2023}
}

@article{zhang2024earthgpt,
  title={{EarthGPT}: A Universal Multi-modal Large Language Model for Multi-sensor Image Comprehension in Remote Sensing Domain},
  author={Zhang, Wei and Cai, Miaoxin and Zhang, Tong and others},
  journal={IEEE Transactions on Geoscience and Remote Sensing},
  year={2024},
  publisher={IEEE}
}

@article{xie2024wildfiregpt,
  title={{WildfireGPT}: Tailored Large Language Model for Wildfire Analysis},
  author={Xie, Yangxinyu and Jiang, Bowen and Mallick, Tanwi and Bergerson, Joshua David and Hutchison, John K and Verner, Duane R and Branham, Jordan and Alexander, M Ross and Ross, Robert B and Feng, Yan and Levy, Leslie-Anne and Su, Weijie and Taylor, Camillo J},
  journal={arXiv preprint arXiv:2402.07877},
  year={2024}
}

@inproceedings{zheng2023judging,
  title={Judging {LLM}-as-a-Judge with {MT-Bench} and Chatbot Arena},
  author={Zheng, Lianmin and Chiang, Wei-Lin and Sheng, Ying and Zhuang, Siyuan and Wu, Zhanghao and Zhuang, Yonghao and Lin, Zi and Li, Zhuohan and Li, Dacheng and Xing, Eric P and Zhang, Hao and Gonzalez, Joseph E and Stoica, Ion},
  booktitle={Advances in Neural Information Processing Systems (NeurIPS)},
  volume={36},
  year={2023}
}

@article{li2024llm_judges_survey,
  title={{LLMs}-as-Judges: A Comprehensive Survey on {LLM}-based Evaluation Methods},
  author={Li, Shuang and Ye, Jianye and others},
  journal={arXiv preprint arXiv:2412.05579},
  year={2024}
}

@article{ayanzadeh2024floorplan2guide,
  title={{Floorplan2Guide}: {LLM}-Guided Floorplan Parsing for {BLV} Indoor Navigation},
  author={Ayanzadeh, Aydin and Oates, Tim},
  journal={arXiv preprint arXiv:2412.18120},
  year={2024}
}

\end{document}